\documentclass{article}

\usepackage{graphicx}
\usepackage{microtype}      % microtypography
\usepackage{subfigure}

\usepackage{arxiv2}

\usepackage[utf8]{inputenc} % allow utf-8 input
\usepackage[T1]{fontenc}    % use 8-bit T1 fonts
\usepackage{hyperref}       % hyperlinks
\usepackage{url}            % simple URL typesetting
\usepackage{booktabs}       % professional-quality tables
\usepackage{amsfonts}       % blackboard math symbols
\usepackage{nicefrac}       % compact symbols for 1/2, etc.

\usepackage{multirow}

\newcommand{\ssymbol}[1]{^{\@fnsymbol{#1}}}

\def\@fnsymbol#1{\ensuremath{\ifcase#1\or *\or \dagger\or \ddagger\or
   \mathsection\or \mathparagraph\or \|\or **\or \dagger\dagger
   \or \ddagger\ddagger \else\@ctrerr\fi}}

\title{MIMIC-CXR-JPG, a large publicly available database of labeled chest radiographs}

\begin{document}

\maketitle

Alistair E.~W.~Johnson\textsuperscript{1$*$},
Tom J.~Pollard\textsuperscript{1},
Nathaniel R.~Greenbaum\textsuperscript{3},
Matthew P.~Lungren\textsuperscript{4},
Chih-ying~Deng\textsuperscript{5}, \\
Yifan Peng\textsuperscript{6},
Zhiyong Lu\textsuperscript{6},
Roger G.~Mark\textsuperscript{1},
Seth J.~Berkowitz\textsuperscript{2},
Steven Horng\textsuperscript{3}
\vspace{0.3cm}
\\
\textsuperscript{1} Institute of Medical Engineering \& Science, Massachusetts Institute of Technology, Cambridge, MA, USA
\\
\textsuperscript{2} Department of Radiology, Beth Israel Deaconess Medical Center, Boston, MA, USA
\\
\textsuperscript{3} Department of Emergency Medicine, Beth Israel Deaconess Medical Center, Boston, MA, USA
\\
\textsuperscript{4} Department of Radiology, Stanford University, Palo Alto, CA, USA
\\
\textsuperscript{5} Department of Biomedical Informatics, Harvard Medical School, Boston, MA, USA
\\
\textsuperscript{6} National Center for Biotechnology Information, National Library of Medicine, National Institutes of Health, USA
\vspace{0.3cm}
\\
$*$ Correspondence to: aewj@mit.edu
\vspace{0.3cm}
\newline
\begin{abstract}

Chest radiography is an extremely powerful imaging modality, allowing for a detailed inspection of a patient's thorax, but requiring specialized training for proper interpretation.
With the advent of high performance general purpose computer vision algorithms, the accurate automated analysis of chest radiographs is becoming increasingly of interest to researchers.
However, a key challenge in the development of these techniques is the lack of sufficient data.
Here we describe MIMIC-CXR-JPG v2.0.0, a large dataset of 377,110 chest x-rays associated with 227,827 imaging studies sourced from the Beth Israel Deaconess Medical Center between 2011 - 2016.
Images are provided with 14 labels derived from two natural language processing tools applied to the corresponding free-text radiology reports.
MIMIC-CXR-JPG is derived entirely from the MIMIC-CXR database, and aims to provide a convenient processed version of MIMIC-CXR, as well as to provide a standard reference for data splits and image labels.
All images have been de-identified to protect patient privacy.
The dataset is made freely available to facilitate and encourage a wide range of research in medical computer vision.

\end{abstract}


\section{Introduction}

Chest radiography is a common imaging modality used to assess the thorax and the most common medical imaging study in the world. Chest radiographs are used to identify acute and chronic cardiopulmonary conditions, verify that devices such as pacemakers, central lines, and tubes are correctly positioned, and to assist in related medical workups.
In the U.S., the number of radiologists as a percentage of the physician workforce is decreasing \cite{rosenkrantz2015us}, and the geographic distribution of radiologists favors larger, more urban counties \cite{rosenkrantz2018county}.
Delays and backlogs in timely medical imaging interpretation have demonstrably reduced care quality in such large health organizations as the U.K. National Health Service \cite{rimmer2017radiologist} and the U.S. Department of Veterans Affairs \cite{bastawrous2017improving}.
The situation is even worse in resource-poor areas, where radiology services are extremely scarce. As of 2015, only 11 radiologists served the 12 million people of
Rwanda \cite{rosman2015imaging}, while the entire country of Liberia, with a population of four million, had only two practicing radiologists \cite{ali2015diagnostic}.
Accurate automated analysis of radiographs has the potential to improve the efficiency of radiologist workflow and extend expertise to under-served regions.

The combination of burgeoning datasets with increasingly sophisticated algorithms has resulted in a number of significant advances in other application areas of computer vision \cite{halevy2009unreasonable, sun2017revisiting}. A key requirement in the application of these advances to automated chest radiograph analysis is sufficient data. Over time, progressively larger databases have been made available. The Japanese Society of Radiological Technology (JSRT) Database contains 247 images with labels of chest nodules as confirmed by subsequent computed tomography (CT) \cite{shiraishi2000development}. Notably, the dataset is provided with annotations segmenting the lungs and heart. The Open-I Indiana University Chest X-ray dataset contains 8,121 images associated with 3,996 de-identified radiology reports \cite{demner2015preparing}. More recently, the NIH released ChestX-ray14 (originally ChestX-ray8), a collection of 112,120 frontal chest radiographs from 30,805 distinct patients with 14 binary labels indicating existence pathology or lack of pathology \cite{wang2017chestx}.

The MIMIC-CXR database aimed to galvanize research around automated analysis of chest radiographs. The chest radiographs in MIMIC-CXR are published in DICOM format, which is commonly used in clinical practice. DICOM is a well defined binary file format which stores a large amount of meta-data with the pixel values of the image. Unfortunately, due to the complexity of the application domain (radiology), the DICOM file format can be difficult to comprehend, creating an undesirable barrier for those traditionally outside of the medical domain. Outside of radiology, digital images tend to be stored using one of a number of more common general purpose formats.
One particularly common format, JPEG, achieves significant savings in image storage size using a lossy compression algorithm. While the loss of information is undesirable, the benefits of a reduced image storage size are many and so the JPG image format remains popular among computer vision researchers.

The primary goal of the MIMIC-CXR-JPG database is to provide a standard reference for JPEG images derived from the DICOM files. As DICOMs contain higher pixel depth than can be perceived by the human eye, a design decision must be made in converting the 16-bit depth raw images into 12-bit depth images in JPEG format. Furthermore, a number of image pixel normalization strategies are employed in computer vision, and providing the most common approach as a reference database saves researchers time and makes it easier to compare derivative works.
Images are also provided with one or more labels derived from the corresponding free-text radiology report using open source labelers \cite{peng2018negbio, irvin2019chexpert}. While other researchers can derive structured labels from the free-text radiology reports in MIMIC-CXR, providing labels here ensures their derivation is consistent.

MIMIC-CXR-JPG is de-identified to satisfy the US Health Insurance Portability and Accountability Act of 1996 (HIPAA) Safe Harbor requirements. Protected health information (PHI) has been removed. Randomly generated identifiers are used to group distinct reports and patients.

\section{Chest radiographs}

% x-rays form a study, usually one front and one side
% DICOM format, converted to JPG
% meta-data scrubbed
% images contain burned in annotations
% if these annotations contained PHI, they were removed
%   - burned in annotation flag set to YES
%   - black box drawn around the annotation

Chest radiographs were sourced from the hospital picture archiving and communication system (PACS) in Digital Imaging and Communications in Medicine (DICOM) format. All studies for patients admitted to the emergency department between 2011 - 2016 were queried.
Images were linked to corresponding radiology reports using the hospital's radiology information system.
Images sometimes contain ``burned in'' annotations: areas where pixel values have been modified after image acquisition in order to display text. Annotations contain relevant information including: image orientation, anatomical position of the subject, timestamp of image capture, and so on. The resulting image, with annotations encoded within the pixel themselves, is then transferred from the modality to PACS. Since the annotations are applied at the modality, it is impossible to recover the original image without annotations.
As all patient PHI must be removed to satisfy HIPAA Safe Harbor, images were de-identified using a custom algorithm which removed dates and patient identifiers, but retained radiologically relevant information such as orientation.
The algorithm applied an ensemble of image preprocessing and optical character recognition approaches to detect text within an image.
Text was identified due to its significant contrast with the background, and due to its consistent pixel value within an image.
Suspected PHI was removed by setting all pixel values in a bounding box encompassing the PHI to black. Subsequent to deidentification, we manually reviewed 6,900 radiographs for PHI. Each image was reviewed by two independent annotators. 180 images were identified for a secondary consensus review; none of which ultimately had PHI. The most common causes for annotators to request consensus review were: (1) existence of a support device such as a pacemaker, (2) text identifying in-hospital location (e.g. ``MICU''), and (3) obscure text relating to radiograph technique (e.g. ``prt rr slot 11'').

After de-identification, images were exported in the JPEG standard format.
First, the image pixels were extracted from the DICOM file using the pydicom library \cite{pydicom}. Pixel values were normalized to the range [0, 255] by subtracting the lowest value in the image, dividing by the highest value in the shifted image, truncating values, and converting the result to an unsigned integer.
The DICOM field PhotometricInterpretation was used to determine whether the pixel values were inverted, and if necessary images were inverted such that air in the image appears white (highest pixel value), while the outside of the patient's body appears black (lowest pixel value).
The OpenCV library was then used to histogram equalize the image with the intention of enhancing contrast. Histogram equalization involves shifting pixel values towards 0 or towards 255 such that all pixel values 0 through 255 have approximately equal frequency. Images were then converted to JPEG files using OpenCV with a quality factor of 95.
Pixel data were normalized to the unit interval, and bit-depth was subsequently scaled to 8-bit (0-255).
If necessary, image intensity values were inverted to the ensure the image transitioned from dark to bright as pixel value increased. Histogram equalization was then applied, and the image was written out in the compressed JPEG format with a quality value of 95.

\begin{figure}[!htb]
% files/p10/p10000032/s50414267/02aa804e-bde0afdd-112c0b34-7bc16630-4e384014
\centering
  \includegraphics[width=0.8\linewidth]{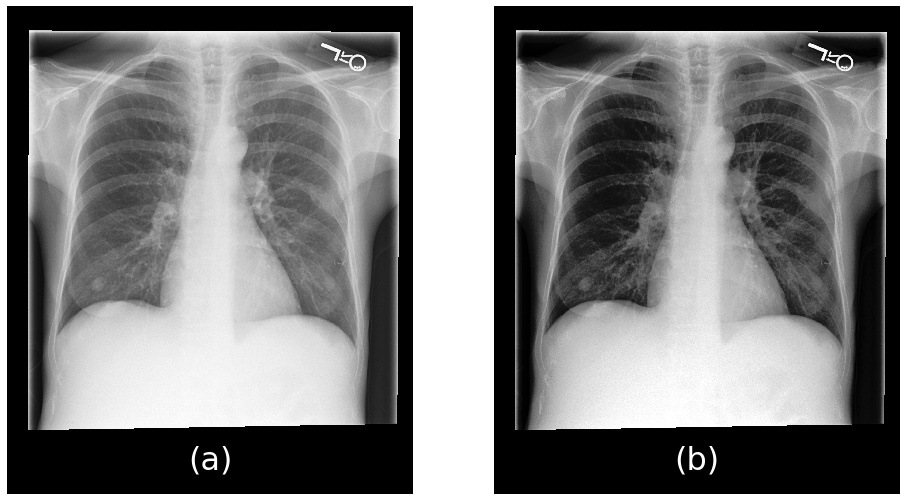}
  \caption{Example of an image converted from DICOM (a) to JPEG (b).}
  \label{fig:jpg_conversion}
\end{figure}

Note that, aside from de-identification and conversion to JPEG, no filtering or processing of the images was performed.
Figure \ref{fig:jpg_conversion} provides a comparison of an image read directly from the DICOM and the histogram equalized JPEG format file. The default parameters were used to display the DICOM image.

\section{Labeling of the reports}

Radiology reports at the source hospital are semi-structured, with radiologists documenting their interpretations in titled sections. The structure of these reports are generally consistent through the use of standardized documentation templates, though can drift over time as the template changed. There can also be some inter-reporter variability as the structure of the reports are not enforced by the user interface and can be overridden by the user.
The two primary sections of interest are \emph{findings}; a natural language description of the important aspects in the image, and \emph{impression}; a short summary of the most immediately relevant findings.
Labels for the images were derived from either the impression section, the findings section (if impression was not present), or the final section of the report (if neither impression nor findings sections were present). Of the total 227,835 reports, 189,561 (83.2\%) had an impression section, 27,684 (12.2\%) had a findings section, and 10,514 (4.6\%) had an equivalent section not explicitly labeled as findings or impression. 8 of the reports did not have text for labeling.

Labels were derived using two open source labeler tools; NegBio and CheXpert \cite{peng2018negbio, irvin2019chexpert}.
NegBio is an open-source rule based tool for negation and uncertain detection in radiology reports. NegBio takes as input a sentence with pre-tagged mentions of medical findings, and determines whether a specific finding is negative or uncertain. More detail is provided in the NegBio article \cite{peng2018negbio}.
CheXpert is a rule based classifier which proceeds in three stages: (1) extraction, (2) classification, and (3) aggregation. In the extraction stage, all mentions of a label are identified, including alternate spellings, synonyms, and abbreviations (e.g. for pneumothorax, the words ``pneumothoraces'' and ``ptx'' would also be captured). Mentions are then classified as positive, uncertain, or negative using local context.
Finally, aggregation is necessary as there may be multiple mentions of a label. More detail is provided in the CheXpert article \cite{irvin2019chexpert}.

In the labeling of reports for MIMIC-CXR-JPG, the mention patterns defined in CheXpert were used for NegBio. As such, the categories overlap with the CheXpert dataset \cite{irvin2019chexpert}, but not the ChestX-ray 14 dataset \cite{wang2017chestx}.
Example reports with labels are shown in Table~\ref{tab:ex_reports}. Table~\ref{tab:stats_reports} shows the frequency of various labels in the reports in the majority subset of the images.
A fourth category, ``Disagreement'', highlights instances where the CheXpert and NegBio tools disagreed on the label.
 
\begin{table}[ht!]
\centering
 \caption{Example radiology reports with labels. Labels in italics are \emph{negated}. Labels with a \textsuperscript{u} are uncertain.}
  \begin{tabular}{lp{9cm}l}
    \toprule
    \textbf{Section}     & \textbf{Report}     & \textbf{Label}\\
    \midrule
    Impression & No evidence of acute cardiopulmonary process.  & No Finding     \\ % index 258
    \midrule
    Findings   & \multirow{7}{9cm}{The left lung is relatively well aerated and clear.  The right
 hemithorax is markedly opacified with volume loss, circumferential pleural
 thickening and pleural fluid with near complete opacification of the right
 lung with right basal pleural catheter noted. Hydropneumothorax previously
 seen is not as well evaluated on this not fully upright film. Cardiac contours
 are somewhat obscured but unremarkable.}  & \emph{No Cardiomegaly} \\ 
 & & \emph{No Enlarged Cardiomediastinum} \\
 & & Pneumothorax\textsuperscript{u} \\ 
 & & Airspace Opacity \\ 
 & & Pleural Effusion \\ 
 & & Pleural Other \\ 
 & & Support Devices  \\ % index 4028
 \midrule
    Other      & \multirow{7}{9cm}{Cardiac size is top normal.
Bibasilar opacities, larger on the left side,
could be due to atelectasis but superimposed infection cannot be excluded.
If any, there is a small right pleural effusion.  There is elevation of the right
hemidiaphragm.  There is mild vascular congestion.}
    & \emph{No Cardiomegaly} \\ 
 & & Atelectasis\textsuperscript{u} \\ 
 & & Pneumonia\textsuperscript{u} \\ 
 & & Edema\textsuperscript{u} \\ 
 & & Airspace Opacity \\ 
 & & Pleural Effusion \\ % 50228105
    \bottomrule
  \end{tabular}
  \label{tab:ex_reports}
\end{table}

% addendums are not necessarily used

% table X: frequency of labels in the final dataset
\begin{table}[ht!]
\centering
  \caption{Frequency of labels in MIMIC-CXR-JPG on the training subset of 222,750 unique radiologic studies (8 studies were not labeled).}
\begin{tabular}{lrrrr}
\toprule
{} &        Positive &          Negative &        Uncertain &  Disagreement \\
\midrule
Atelectasis                &  45,088 (19.8\%) &      937.0 (0.4\%) &   9,897.0 (4.3\%) &  1,744 (0.8\%) \\
Cardiomegaly               &  39,094 (17.2\%) &   15,860.0 (7.0\%) &   5,924.0 (2.6\%) &  5,924 (2.6\%) \\
Consolidation              &   10,487 (4.6\%) &    7,939.0 (3.5\%) &   3,022.0 (1.3\%) &  1,628 (0.7\%) \\
Edema                      &  26,455 (11.6\%) &  25,246.0 (11.1\%) &  11,781.0 (5.2\%) &  2,351 (1.0\%) \\
Enlarged Cardiomediastinum &    7,004 (3.1\%) &    5,271.0 (2.3\%) &   9,307.0 (4.1\%) &    255 (0.1\%) \\
Fracture                   &    3,768 (1.7\%) &      880.0 (0.4\%) &     299.0 (0.1\%) &    884 (0.4\%) \\
Lung Lesion                &    6,129 (2.7\%) &      842.0 (0.4\%) &   1,020.0 (0.4\%) &    296 (0.1\%) \\
Lung Opacity               &  50,916 (22.3\%) &    2,868.0 (1.3\%) &   2,110.0 (0.9\%) &  2,531 (1.1\%) \\
No Finding                 &  75,163 (33.0\%) &        - &       - &  3,906 (1.7\%) \\
Pleural Effusion           &  53,188 (23.3\%) &  27,072.0 (11.9\%) &   5,345.0 (2.3\%) &  1,667 (0.7\%) \\
Pleural Other              &    1,961 (0.9\%) &      120.0 (0.1\%) &     728.0 (0.3\%) &     93 (0.0\%) \\
Pneumonia                  &   15,769 (6.9\%) &  24,205.0 (10.6\%) &  17,789.0 (7.8\%) &  1,422 (0.6\%) \\
Pneumothorax               &    9,317 (4.1\%) &  42,335.0 (18.6\%) &     868.0 (0.4\%) &  1,328 (0.6\%) \\
Support Devices            &  65,637 (28.8\%) &    3,070.0 (1.3\%) &      96.0 (0.0\%) &  1,831 (0.8\%) \\
\bottomrule
\end{tabular}
  \label{tab:stats_reports}
\end{table}

\section{Training, validation, and test sets}

To ensure consistent evaluation of models, we have organized the data into training, validation, and test sets.
The test set contains all studies for patients who had at least one report labelled in our manual review. 
We are not publicly releasing the test set.
The validation set contains a random set of 500 patients and all of their associated studies.
This set is made publicly available in a separate `valid' folder.
Finally, all remaining studies are made available in the training set.
Table~\ref{tab:train_val_test} provides summary information for studies in the three datasets.
Note the enrichment of findings in the test set caused by the stratified sampling done to ensure sufficient coverage of all pathologies.

\begin{table}[ht!]
\centering
 \caption{Summary of the images split into training, validation, and test sets.}
 \begin{tabular}{lccc}
\toprule
Dataset &           Train &      Validate &          Test \\
\midrule
Number of images   &          368960 &          2991 &          5159 \\
\ \ Frontal            &  248020 (67.2\%) &  2041 (68.2\%) &  3653 (70.8\%) \\
\ \ Lateral            &  120795 (32.7\%) &   949 (31.7\%) &  1502 (29.1\%) \\
\ \ Other              &      145 (0.0\%) &      1 (0.0\%) &      4 (0.1\%) \\
\midrule
Number of studies  &          222758 &          1808 &          3269 \\
\ \ with a finding   &  170420 (76.5\%) &  1394 (77.1\%) &  2912 (89.1\%) \\
\midrule
Number of patients &           64586 &           500 &           293 \\
\ \ with a finding   &   44157 (68.4\%) &   344 (68.8\%) &   288 (98.3\%) \\
\bottomrule
\end{tabular}
  \label{tab:train_val_test}
\end{table}

\section{Validation of labels}

A random set of reports were selected for validation.
Stratified sampling was used to ensure adequate capture of the various pathologies.
A total of 687 reports were reviewed by a board certified radiologist with 8 years experience (ML) and manually labeled according to the 14 categories in CheXpert.
The labeling process followed guidelines set forth by the authors of the CheXpert labeler and described therein \cite{irvin2019chexpert}.

The two label algorithms were evaluated in three tasks: mention extraction, negation detection, and uncertainty detection. For the mention extraction task, any assigned label (positive, negative, or uncertain) is considered a positive prediction, while blank (no mention) is considered a negative prediction. For negation detection, negated labels are positive while all other labels are negative. 
Finally, for uncertainty detection, uncertain labels are positive while all other labels are negative.
The harmonic mean of the sensitivity and positive predictive value, referred to as the F1 score, was calculated for each group independently. Table~\ref{tab:mention} lists the performance for the mentioning of a label, Table~\ref{tab:uncertainty} lists the performance for uncertainty classification, and Table~\ref{tab:negation} lists the performance for negation classification.

\begin{table}[ht!]
\centering
 \caption{Evaluation of the CheXpert mention patterns on 687 manually labeled reports. The aim is to detect any utterance of the finding, regardless of uncertainty. The CheXpert mention patterns were used for both NegBio and CheXpert in this work.}
\begin{tabular}{lrrrr}
\toprule
Finding &  Precision &    Recall &        F1 &  Number of positive cases \\
\midrule
No Finding                 &   0.382 &  0.867 &  0.531 &         30 \\
Enlarged Cardiomediastinum &   0.375 &  0.600 &  0.462 &         70 \\
Cardiomegaly               &   0.814 &  0.910 &  0.859 &        235 \\
Lung Lesion                &   0.861 &  0.848 &  0.855 &         66 \\
Lung Opacity               &   0.715 &  0.907 &  0.800 &        194 \\
Edema                      &   0.799 &  1.000 &  0.888 &        227 \\
Consolidation              &   0.886 &  0.979 &  0.930 &         95 \\
Pneumonia                  &   0.928 &  0.986 &  0.956 &        223 \\
Atelectasis                &   0.893 &  1.000 &  0.944 &        218 \\
Pneumothorax               &   0.945 &  0.995 &  0.970 &        226 \\
Pleural Effusion           &   0.968 &  0.978 &  0.973 &        370 \\
Pleural Other              &   0.477 &  0.778 &  0.592 &         27 \\
Fracture                   &   0.870 &  0.940 &  0.904 &         50 \\
Support Devices            &   0.823 &  0.974 &  0.892 &        234 \\
\bottomrule
\end{tabular}
  \label{tab:mention}
\end{table}

\begin{table}[ht!]
\centering
 \caption{Evaluation of the uncertainty patterns on 687 manually labeled reports. Uncertainty was not evaluated for ``No Finding'', ``Lung Opacity'', ``Pleural Other'', or ``Support Devices'' as no cases of uncertainty were labeled. A value of ``-'' indicates the measure could not be calculated.}
 \begin{tabular}{lrrrrrrr}
\toprule
 & \multicolumn{2}{l}{Precision} & \multicolumn{2}{l}{Recall} & \multicolumn{2}{l}{F1} & Positive cases \\
Uncertainty &    NegBio & CheXpert & NegBio & CheXpert & NegBio & CheXpert &     \\
\midrule
% No Finding                 &       NaN &      NaN &    NaN &      NaN &    NaN &      NaN &         0 \\
Enlarged Cardiomediastinum &     0.033 &    0.036 &  1.000 &    1.000 &  0.065 &    0.069 &         1 \\
Cardiomegaly               &     0.156 &    0     &  0.500 &    0     &  0.237 &      - &        14 \\
Lung Lesion                &     0     &    0     &  0     &    0     &    - &      - &         8 \\
% Lung Opacity               &       NaN &      NaN &    NaN &      NaN &    NaN &      NaN &         0 \\
Edema                      &     0.102 &    0.125 &  0.500 &    0.600 &  0.169 &    0.207 &        10 \\
Consolidation              &     0.529 &    0.273 &  0.529 &    0.176 &  0.529 &    0.214 &        17 \\
Pneumonia                  &     0.432 &    0.407 &  0.613 &    0.565 &  0.507 &    0.473 &        62 \\
Atelectasis                &     0.333 &    0.289 &  0.706 &    0.647 &  0.453 &    0.400 &        17 \\
Pneumothorax               &     0.375 &    0.250 &  0.375 &    0.125 &  0.375 &    0.167 &         8 \\
Pleural Effusion           &     0.432 &    0.414 &  0.889 &    0.667 &  0.582 &    0.511 &        18 \\
% Pleural Other              &       NaN &      NaN &    NaN &      NaN &    NaN &      NaN &         0 \\
Fracture                   &     0.500 &    0     &  0.500 &    0     &  0.500 &      - &         2 \\
% Support Devices            &       NaN &      NaN &    NaN &      NaN &    NaN &      NaN &         0 \\
\bottomrule
\end{tabular}

  \label{tab:uncertainty}
\end{table}

\begin{table}[ht!]
\centering
 \caption{Evaluation of the negation patterns on 687 manually labeled reports. Negation was not evaluated for ``No Finding'' as no cases of negation were labeled. A value of ``-'' indicates the measure could not be calculated.}
\begin{tabular}{lrrrrrrr}
\toprule
{} & \multicolumn{2}{l}{Precision} & \multicolumn{2}{l}{Recall} & \multicolumn{2}{l}{F1} & Positive cases \\
Negation &    NegBio & CheXpert & NegBio & CheXpert & NegBio & CheXpert &     \\
\midrule
% No Finding                 &       NaN &      NaN &    NaN &      NaN &    NaN &      NaN &         0 \\
Enlarged Cardiomediastinum &     0.654 &    0.654 &  0.607 &    0.607 &  0.630 &    0.630 &        28 \\
Cardiomegaly               &     0.855 &    0.855 &  0.720 &    0.720 &  0.781 &    0.781 &        82 \\
Lung Lesion                &     0.500 &    0.500 &  0.500 &    0.500 &  0.500 &    0.500 &         4 \\
Lung Opacity               &     0.429 &    0.533 &  0.391 &    0.348 &  0.409 &    0.421 &        23 \\
Edema                      &     0.713 &    0.714 &  0.847 &    0.824 &  0.774 &    0.765 &        85 \\
Consolidation              &     0.917 &    0.917 &  0.957 &    0.957 &  0.936 &    0.936 &        23 \\
Pneumonia                  &     0.836 &    0.868 &  0.735 &    0.711 &  0.782 &    0.781 &        83 \\
Atelectasis                &     0.333 &    0.300 &  0.750 &    0.750 &  0.462 &    0.429 &         4 \\
Pneumothorax               &     0.919 &    0.926 &  0.955 &    0.911 &  0.937 &    0.918 &       179 \\
Pleural Effusion           &     0.906 &    0.919 &  0.939 &    0.963 &  0.922 &    0.940 &        82 \\
Pleural Other              &     0     &    0     &  0     &    0     &    - &      - &         2 \\
Fracture                   &     0.600 &    0     &  0.375 &    0     &  0.462 &      - &         8 \\
Support Devices            &     0.200 &    0     &  0.400 &    0     &  0.267 &      - &         5 \\
\bottomrule
\end{tabular}

  \label{tab:negation}
\end{table}

\newpage{}
\section{Data availability}

All data is made available on PhysioNet\footnote{https://www.physionet.org/content/mimic-cxr-jpg/} \cite{mimiccxrjpg, goldberger2000physiobank}.
Use of the dataset is free to all researchers after signing of a data use agreement which stipulates, among other items, that (1) the user will not share the data, (2) the user will make no attempt to reidentify individuals, and (3) any publication which makes use of the data will also make the relevant code available.

MIMIC-CXR-JPG is wholly derived from MIMIC-CXR\footnote{https://www.physionet.org/content/mimic-cxr/} \cite{mimiccxr}.
The source data, MIMIC-CXR, contains the same images in DICOM format with the free-text radiology reports which were the source of the labels.
Due to the sensitivity of this dataset, access will require completion of a training course in human subjects research, as is the process for MIMIC-III \cite{johnson2016mimic} and eICU-CRD \cite{pollard2018eicu}.

Code used to generate MIMIC-CXR-JPG and the summaries in this paper has been made publicly available\footnote{https://github.com/MIT-LCP/mimic-cxr/} \citep{mimic-cxr-code}.

\section{Conclusions}

MIMIC-CXR-JPG is a large, publicly available dataset of chest radiographs from over 220,000 studies performed at the BIDMC. The dataset contains labels for a number of common pathologies and will provide a benchmark for a number of medically relevant computer vision tasks.

\section*{Acknowledgements}

We would like to acknowledge the Stanford Machine Learning Group and the Stanford AIMI center for their help in running the chexpert labeler and for their insight into the work; in particular we would like to thank Jeremy Irvin and Pranav Rajpurkar.
We would also like to acknowledge the BIDMC for their continued collaboration.

This work was supported by grant NIH-R01-EB017205 from the National Institutes of Health.
This work was also supported by the Intramural Research Programs and grant K99LM013001 of the NIH National Library of Medicine.

The MIT Laboratory for Computational Physiology received funding from Philips Healthcare to create the database described in this paper.

\newpage{}
\bibliographystyle{vancouver}
\bibliography{references}

\end{document}